\documentclass{article}
\usepackage{proceed2e}
\usepackage{fullname}
\usepackage{haldefs}
\usepackage{algorithm}
\usepackage{algorithmic}
\usepackage{psfig}
\usepackage{url}
\usepackage{multirow}

\title{Fast search for Dirichlet process mixture models}
 
\author{ {\bf Hal Daum\'e III} \\  
School of Computing \\  
University of Utah \\ 
Salt Lake City, UT 84112 \\ 
}

\begin{document}
\maketitle

\begin{abstract}
Dirichlet process (DP) mixture models provide a flexible Bayesian
framework for density estimation.  Unfortunately, their flexibility
comes at a cost: inference in DP mixture models is computationally
expensive, even when conjugate distributions are used.  In the common
case when one seeks only a maximum a posteriori assignment of data
points to clusters, we show that search algorithms provide a practical
alternative to expensive MCMC and variational techniques.  When a true
posterior sample is desired, the solution found by search can serve as
a good initializer for MCMC.  Experimental results show that using
these techniques is it possible to apply DP mixture models to very
large data sets.
\end{abstract}

\section{INTRODUCTION}

Dirichlet process (DP) mixture models provide a flexible Bayesian
solution to nonparametric density estimation.  Their flexibility
derives from the fact that one need not specify a number of mixture
components a priori.  In practice, DP mixture models have been used
for problems in genomics \cite{xing04haplotype}, relational learning
\cite{xu05dirichlet}, data mining \cite{daume05dpscm} and vision
\cite{sudderth05visual}.  Despite these successes, the flexibility of
DP mixture models comes at a high computational cost.  Standard
algorithms based on MCMC, such as those described by
\namecite{neal98dpmm}, are computationally expensive and can take a
long time to converge to the stationary distribution.  Variational
techniques \cite{blei05variational} are an attractive alternative, but
are difficult to implement and can remain slow.

In this paper, we show that standard search algorithms, such as A*,
and beam search, provide an attractive alternative to these expensive
techniques.  Our algorithms allows one to apply DP mixture models to
very large data sets.  Like variational approaches to DP mixture
models, we focus on conjugate distributions from the exponential
family.  Unlike MCMC techniques, which can produce samples of cluster
assignments from the corresponding posterior, our search-based
techniques will only find an approximate MAP cluster assignment.  We
do not believe this to be a strong limitation: in practice, the
applications cited above all use MCMC techniques to draw a sample and
then simply choose from this sample the single assignment with the
highest posterior probability.  If one needs samples from the
posterior, then the solution found by our methods could initialize
MCMC.

\section{DIRICHLET PROCESSES}

The Dirichlet process, introduced by \namecite{ferguson73bayesian}, is
a distribution over distributions.  The DP is parameterized by a base
measure $G_0$ and a concentration parameter $\al$.  We write $G \|
G_0, \al \by \DP(G_0,\al)$ for a draw of a distribution $G$ from the
Dirichlet process.  We may then draw parameters $\th_{1:N} \| G \by
G$.  By marginalizing over $G$, we find that the draws of the
parameters $\th$ obey a P\`olya urn scheme
\cite{blackwell-macqueen73polya}: previously drawn values of $\th$
have strictly positive probability of being redrawn, thus making the
underlying probability measure discrete (with probability one).

By using a Dirichlet process at the top of a hierarchical model, one
obtains a Dirichlet process mixture model \cite{antoniak74mixtures}.
Here, one treats the $n$th parameter $\th_n$ as being associated with
the $n$th observation, using some likelihood function $F$.  This
yields the mixture model shown in Eq~\eqref{eq:dpmm}.

\begin{hierarchical} \label{eq:dpmm}
G     & \al, G_0 & \DP(\al, G_0) \\
\th_n & G        & G \\
x_n   & \th_n    & F(\th_n)
\end{hierarchical}

The clustering property of the DP prefers that fewer than $N$ distinct
values of $\th$ will be used.  If $K \leq N$ values are used, the
$x_n$s can be seen to be clustered into one of $K$ clusters,
determined by the $\th$ values.

\subsection{PROPERTIES}

DP mixture models posses several properties that are useful for
further analysis.  The most basic is their well-known property of
exchangeability: samples from the process are order independent.  This
leads to efficient MCMC techniques (see Section~\ref{sec:mcmc}).  An
additional property of the DP that will be useful for our analysis is
given as Proposition 3 by \namecite{antoniak74mixtures}.  He gives an
explicit form for the probability of individual clusterings.  In
particular, suppose that a sample $x_1, \dots, x_N$ is drawn from a DP
mixture model as in Eq~\eqref{eq:dpmm}.  At most $N$ distinct values
of $\th$ will be used (and typically far fewer).  Define a vector
$m_1, \dots, m_I$ by: $m_i$ is the number of $\th$s that appear
exactly $i$ times.  Thus, $N = \sum_i i m_i$ and $\sum_i m_i$ is the
total number of clusters.  \namecite{antoniak74mixtures} gives an
explicit formulation for the distribution of counts $\vec m$:

\begin{equation} \label{eq:m}
P(\vec m \| \al, N) =
  \frac {N!} {\al^{(N)}}
  \frac {\al^{\sum_{i=1}^I m_i}}  {\prod_{i=1}^I i^{m_i} (m_i!)}
\end{equation}

\noindent
where $\al^{(N)}$ denotes the rising factorial function: $\al^{(0)} =
1$ and $\al^{(n)} = \al (\al+1) \cdots (\al+n-1)$.

\subsection{MCMC TECHNIQUES} \label{sec:mcmc}

When the mean distribution $G_0$ of the DP is conjugate to the
likelihood function $F$, one can analytically integrate out $G$ and
the $\th$s from Eq~\eqref{eq:dpmm} and only maintain a vector of
cluster assignments, $\vec c_{1:N}$.  The vector $\vec c$ serves to
specify which $x_n$s were generated from the same mixture component
$\th$, so that $x_n$ is drawn according to $F(\th_{c_n})$.  Using the
exchangeability of the DP, a single Gibbs iteration proceeds as
follows \cite[Algorithm 3]{neal98dpmm}.  For each $n$, we draw $c_n$
to be a \emph{new} cluster with probability proportional to $\alpha
H(x_n)$ and draw it equal to an existing cluster $d$ with probability
$N_{-n,d} H(x_n \| \{ x_i \| c_i = d, i \neq n \})$.  Here, $\alpha$
is the concentration parameter for the DP, $N_{-n,d}$ is the number of
elements of $\vec c$ (other than $n$ itself) that are equal to $d$.
Finally, $H(x \| S)$ is the posterior probability of $x$ given that
$S$ has been observed, Eq~\eqref{eq:H}.

\begin{align} \label{eq:H}
H(x \| S) &\defeq \int \ud G_0(\th \| S) F(x \| \th) \\
&= \frac{ \int \ud G_0(\th) F(x \| \th) \prod_{y \in S} F(y \| \th) }
        { \int \ud G_0(\th) \prod_{y \in S} F(y \| \th) }
   \nonumber
\end{align}

By analogy, we will also write $H(\vec x_{1:N} \| S)$ to denote the
marginal posterior probability of $N$ data points.  In general, when
working in the exponential family, $H$ will be available in closed
form and thus the Gibbs sampling is efficient.

\namecite{jain04splitmerge} propose a Metropolis-Hastings sampler
based on splitting and merging existing clusters.  This algorithm is
shown to mix faster than the vanilla Gibbs sampler outlined above.  It
works by first randomly choosing two data points.  If the data points
are currently in different clusters, a proposal is created that merges
the two clusters.  If the data points are currently in the same
cluster, a proposal is generated the splits the cluster.
\namecite{jain04splitmerge} present three variants on this idea.  In
the first variant, splits are determined completely randomly.  In the
second variant, a small Gibbs sampler is run to determine the splits.
In the third variant, a Gibbs sampler is run to determine the splits
\emph{and} the Metropolis-Hastings iterations are interleaved with the
standard Gibbs iterations described above.

\section{SEARCH}

MCMC is an attractive technique for inference in DP mixture models.
However, in many real-world cases, one does not actually need a sample
of cluster assignments from the posterior, but actually seeks only a
single cluster assignment.  In these cases, a sample is extracted from
an MCMC run and a single cluster assignment---the MAP assignment---is
extracted from the sample.  This raises the question: is Gibbs
sampling a good search algorithm?  We show experimentally (see
Section~\ref{sec:artificial}) that it is often not.

\begin{figure}[t]
\framebox{\begin{minipage}[t]{8.1cm}
{\bf \textsf{function}} DPSearch\\
{\bf \textsf{input:}} a scoring function $g$, beam size $b$, data $\vec x_{1:N}$\\
{\bf \textsf{output:}} a clustering $\vec c$
\begin{algorithmic}[1]
\STATE initialize max-queue: $Q \leftarrow [ \langle\rangle ]$
\WHILE{$Q$ is not empty}
\STATE remove state $\vec c_{1:N^0}$ from the front of $Q$
\STATE{\bf \textsf{if}} $N^0 = N$ {\bf \textsf{then return}} $\vec c$
\FORALL{clusters $d$ in $\vec c$ and a new cluster}
\STATE let $\vec c^0 = \vec c \oplus \langle d \rangle$
\STATE compute the score $s = g(\vec c^0, \vec x)$
\STATE update queue: $Q \leftarrow$ Enqueue($Q$, $\vec c^0$, $s$)
\ENDFOR
\IF{$b$ $< \infty$ and $\card{Q} > b$}
\STATE Shrink queue: $Q \leftarrow Q_{1:b}$
\STATE \quad\quad\emph{(drop lowest-scoring elements)}
\ENDIF
\ENDWHILE
\end{algorithmic}
\end{minipage}}
\label{fig:search}
\caption{The generic DP search algorithm.}
\end{figure}

In general, it will be intractable to find the exact MAP solution, as
doing so is NP-hard (by reduction to graph partitioning).  Here, we
describe a set of possible search algorithms one can apply to DP
mixture models for finding the \emph{true} MAP solution for small
problems or an \emph{approximate} MAP solution for large problems.
The generic search algorithm we use is shown in
Figure~\ref{fig:search}.  It takes an ordered set of data points $\vec
x_{1:N}$, a scoring function and a maximum beam size.  It maintains a
max-queue of clusterings of prefixes.  In each iteration, it removes
the most promising element $\vec c$ from the queue and expands it by a
single data point.  The two elements of variability in the algorithm
are the scoring function and the maximum beam size.

The search algorithm is guaranteed to find the maximum a posteriori
clustering if the beam size is $\infty$ and the scoring function is
\emph{admissible}.  In words, $g$ should \emph{over-estimate} the
probability of the \emph{best possible} clustering $\vec c$ that
agrees with $\vec c^0$ on the first $N^0 = \card{\vec c^0}$ elements.
In equations, we write $\vec c \restricted N^0$ to denote the
\emph{restriction} of $\vec c$ to the first $N^0$ elements.  Thus:

\begin{equation}
g(\vec c^0,\vec x) \geq \max_{\vec c : \vec c \restricted N^0=\vec c^0} p(\vec c,\vec x)
\end{equation}

We further require that when $\card{\vec c} = \card{\vec x}$,
equality holds.

However, for efficiency, it is useful to use scoring functions $g$
that occasionally \emph{underestimate} the true posterior probability.
While these functions no longer guarantee that the exact MAP solution
will be found, they are often more efficient because $g$ can be
tighter, even if it is not a strict upper bound (see
Section~\ref{sec:artificial} for supporting evidence).

The posterior probability $p(\vec c,\vec x)$ can be factored as
$p(\vec c) p(\vec x \| \vec c)$.  Here, the probability of a cluster
vector $p(\vec c)$ is given in Eq~\eqref{eq:m} (the mapping from the
$\vec c$ vector to the $\vec m$ vector is straightforward).  The
probability of the data given the clusters is given in
Eq~\eqref{eq:data}.

\begin{equation} \label{eq:data}
p(\vec x \| \vec c) =
  \prod_{d \in \vec c}
    H( \vec x_{\vec c=d} )
\end{equation}

Here, we write $\vec x_{\vec c=d}$ as shorthand for $\{ x_n : c_n = d
\}$.  Our goal is a function $g(\vec c^0, \vec x)$ that upper bounds
the probability of the best clustering $\vec c$ that completes $\vec
c^0$, as in Eq~\eqref{eq:data}.  For brevity, we write $N^0$ to be the
length of $\vec c^0$ and $K^0$ to be the number of clusters in $\vec
c^0$.

An upper bound can be obtained by independently upper-bounding the two
terms, $p(\vec c)$ and $p(\vec x \| \vec c)$.  In fact, we do not
upper bound $p(\vec c)$ but rather explicitly maximize it.  The
algorithm for this computation is given in Section~\ref{sec:c}.  The
maximization of $p(\vec x \| \vec c)$ is more complex and cannot be
performed explicitly.  We give three techniques for this maximization.
The first, trivial computation, is admissible but very loose
(Section~\ref{sec:trivial}).  The second is tighter but still
admissible (Section~\ref{sec:tighter}).  The third is tighter yet, but
is no longer admissible (Section~\ref{sec:inadmissible}).

Matlab code for solving these problems is available online at
\url{http://hal3.name/DPsearch/}.

\subsection{MAXIMIZING THE CLUSTERS} \label{sec:c}

It is possible to explicitly compute the clustering $\vec c$,
beginning with $\vec c^0$, that maximizes the posterior cluster
probability given in Eq~\eqref{eq:m}.  Consider the case of adding a
single data point to $\vec m$.  (Recall that $m_i$ denotes the number
of clusters that contain exactly $i$ data points.)  If this new data
point corresponds to a new cluster, then $m_1$ will increase by one.
If this data point corresponds to an existing cluster that already
contains $\ell$ data points, then $m_\ell$ will decrease by one and
$m_{\ell+1}$ will increase by one.  This gives a \emph{change} in
probability for adding a new data point in Eq~\eqref{eq:m-change}.

\begin{equation} \label{eq:m-change}
\rightarrow \textrm{new}: 
  \frac \al {m_1 + 1}  ~~~~~
\rightarrow \ell: 
  \frac \ell {\ell+1} \frac {m_\ell} {m_{\ell+1}+1}
\end{equation}

By the exchangeability of the underlying process, we obtain
exchangeability on the $\vec m$ vector.  Thus we can \emph{greedily}
search for completions of an initial $\vec m^0$ by executing one of
the two actions in Eq~\eqref{eq:m-change} for the remaining $N-N^0-1$
data points.  In particular, for $N-N^0-1$ steps, we find the value of
$\ell$ (or ``new'') that maximizes Eq~\eqref{eq:m-change} and modify
the corresponding locations in the vector.  After all steps, we simply
compute the probability of the $\vec m$ vector according to
Eq~\eqref{eq:m}.

When $N$ is very large, the loop for computing the optimal $\vec m$
can be time-consuming.  One can greatly accelerate the computation by
noticing that once the largest cluster gets sufficiently large, it
will simply continue to grow for the remainder of the iterations.
Specifically, if in any step the largest cluster is increased from
size $\ell-1$ to $\ell$, and $[\ell m_\ell] /
[(\ell+1)(m_{\ell+1}+1)]$ dominates $\al / [m_1+1]$, then one can stop
the search process and simply further increase $\ell$ with all
remaining elements.  Additionally, it is helpful to cache previously
computed values for repeated use.

\subsection{A TRIVIAL FUNCTION} \label{sec:trivial}

Given a clustering $\vec c^0$ of the first $N^0$ elements, we can
trivially upper bound $p(\vec x \| \vec c)$ from Eq~\eqref{eq:data} by
considering only the first $N^0$ data points.  For this admissible
scoring function, we use Eq~\eqref{eq:trivial}.

\begin{equation} \label{eq:trivial}
g_\textrm{Trivial}(\vec x \| \vec c^0) \defeq
  \prod_{k \in \vec c^0}
    H( \vec x_{\vec c^0 = k} )
\end{equation}

Using the trivial scoring function is essentially equivalent to just
using a path cost with a zero heuristic function in standard A*
search.  As such, we expect this scoring function will lead to an
inefficient search.

\subsection{A TIGHTER FUNCTION} \label{sec:tighter}

The inefficiency of the trivial scoring function given in
Section~\ref{sec:trivial} is due to the fact that it does not take
into account any of the unclustered data points.  We can obtain a
tighter scoring function by accounting for the probability of the as
yet unclustered data points.  We do this by simplifying the
maximization as follows.

\begin{align}
&\hspace{-8mm}\max_{\substack{\vec c :\\\vec c \restricted N^0 = \vec c^0}}
  p(\vec x \| \vec c)\nonumber\\
=&
\max_{\substack{\vec c :\\\vec c \restricted N^0 = \vec c^0}}
\prod_{k \in \vec c}
  H(\vec x_{\vec c=k}) \\
=&
\max_{\substack{\vec c :\\\vec c \restricted N^0 = \vec c^0}}
\prod_{n=1}^N
  H(x_n \| \vec x_{\vec c_{1:n-1}=c_m}) \\
=&
\prod_{n=1}^{N^0}
  H(x_n \| \vec x_{\vec c^0_{1:n-1}=c_m}) \\
&~~~~~~~~~~~~\max_{\substack{\vec c :\\\vec c \restricted N^0 = \vec c^0}}
\prod_{n=N^0+1}^{N}
  H(x_n \| \vec x_{\vec c_{1:n-1}=c_m}) \nonumber\\
\leq& g_{\textrm{Trivial}}(\vec x \| \vec c^0) \\
&\prod_{n=N^0+1}^{N}
  \max_{1 \leq k \leq K^0+1}
  \max_{\substack{\vec c :\\\vec c \restricted N^0 = \vec c^0\\c_n=k}}
  H(x_n \| \vec x_{\vec c_{1:n-1}=c_m}) \nonumber
\end{align}

The key idea for this scoring function is to treat each as-yet
unclustered data point independently.  In particular, for some $n >
N^0$, we know that it will either fall into one of the $K$ clusters
that exist in $\vec c^0$, or it will fall into a new cluster.  So, for
each unclustered point $x_n$, we choose a value $k$ for which cluster
it falls in to.  We then must cluster all remaining points $x_{N^0+1}
\dots x_{n-1}$ as to \emph{only} whether they fall into cluster $k$.

Despite the fact that this latter maximization is simpler, it is still
not tractable (effectively because $H(x \| S)$ is not monotonic in
$S$, even for the exponential family).  The solution we propose is to
replace each remaining $x_m$ ($N^0 < m < n$) with a \emph{replica} of
$x_n$.\footnote{We do not use an actual replica: we use a scaled
replica with norm equal to the maximum norm of \emph{all} $x$.}  In
this way, we \emph{do} obtain monotonicity of $H$ and may simply keep
adding replicas of $x_n$ to $S$ so long as this increases the cluster
probability.\footnote{In certain cases, it is possible to determine
the number of copies that should be added, analytically.  For the
Dirichlet/Multinomial pair, all should be added.  For the
Gaussian/Gaussian pair, one should add sufficiently many copies of
$\be x_n$ (where $\be \geq 1$ is the scaling factor) so as to fully
move the posterior mean to lie at $x_n$ and then add copies of $x_n$
without the scaling factor.}

\subsection{AN INADMISSIBLE FUNCTION} \label{sec:inadmissible}

The foregoing scoring functions are attractive because they provably
lead to optimal clusterings.  However, this optimality comes are a
price: the NP-hardness of the search problem implies that they will
often be inefficient.  This inefficiency comes from their lack of
tightness.  Here, we give a very simple scoring function that is
significantly tighter, and therefore leads to much faster
clusterings.  Unfortunately, this heuristic is no longer admissible
and therefore the search is no longer guaranteed to find the globally
optimal solution.  The function we use is given in Eq~\eqref{eq:ginad}.

\begin{equation} \label{eq:ginad}
g_{\textrm{Inad}}(\vec x \| \vec c^0) \defeq
  g_{\textrm{Trivial}}(\vec x \| \vec c^0)
  \prod_{n=N^0+1}^N H(x_n)
\end{equation}

This scoring function uses the true probability of the existing
clusters and then assigns each new data point to a new cluster.  This
is inadmissible because, for example, if the last two data points are
identical, it would be preferable to cluster them together.

When using this scoring function, the \emph{order} in which the data
points is presented becomes important.  We have found that a useful
heuristic is to order the data points by increasing marginal
likelihood.  This is likely because examples with high marginal
likelihood are more likely to be in their own clusters anyway, and
hence the heuristic is better.  (See Section~\ref{sec:nips} for some
experiments comparing ordering strategies.)

\section{EXPERIMENTAL RESULTS}

We present results on three problems, one based on artificial data,
one based on the MNIST images data set and one based on the NIPS
papers data set.  All experiments were run on a $3.8$GHz Intel Pentium
4 machine with $4$Gb of RAM.

\subsection{ARTIFICIAL DATA} \label{sec:artificial}

\begin{figure*}[t]
\psfig{figure=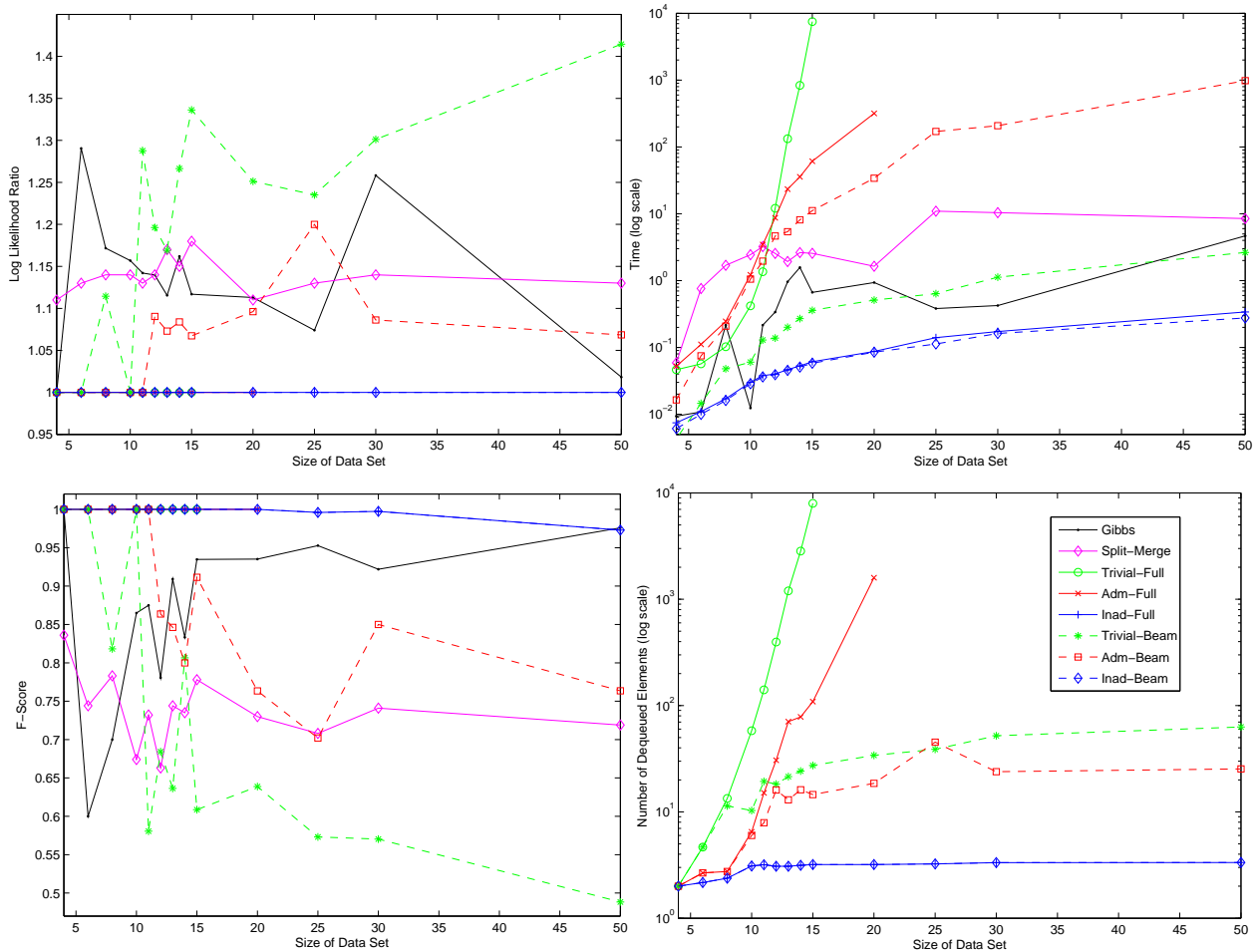,width=17cm}
\caption{Results on artificial data set; x-axis is always the size of
  the data set.  (UL) y-axis is the ration of the data negative log
  likelihood to the data negative log likelihood found by exhaustive
  search (lower is better); (BL) y-axis is f-score (higher is better);
  (UR) y-axis is computation time in seconds (lower is better); (BR)
  y-axis is number of elements enqueued during search (lower is
  better).}
\label{fig:artificial}
\end{figure*}

Our first set of experiments are on artificial data to demonstrate the
scaling properties of the search methods and to compare them directly
to Gibbs sampling.  For these problems, we generate a data set
according to a Gaussian/Gaussian DP mixture model with prior mean zero
and prior variance $10$.  We generate data sets of increasing size ($N
\in \{ 4,6,8,10,11,12,13,14,15,20,25,30,50 \}$).  We first run Gibbs
sampling and split-merge on each data set.  We then run six search
algorithms.  The first three search algorithms run full search using
the three scoring functions described above.  The last three search
algorithms use the three scoring functions described above, but with a
maximum queue size of $10$.  For each data set size, we generate $10$
data sets and average results across these.

Comparison between a sampling approach and a non-sampling approach is
difficult.  We perform our comparison to be as unfair to our proposed
approach as possible: i.e., we try to make sampling look as good as
possible.  We run both samplers as follows.  We do fifteen runs of the
sampler for 1000 iterations each.  The first five are initialized with
a single cluster; the second five are initialized with $N$ clusters;
the last five are initialized randomly with $\log N$ clusters.  For
each run, there will be a sample that achieves the highest log
likelihood.  We choose this iteration as the stopping point.  The best
of the best log likelihoods over the 15 runs is then reported as the
final score.  The \emph{time} reported is the time it took to get to
that log likelihood \emph{in the single run.}  That is, the numbers
reported are overly optimistic in terms of time, and in line with
practice in terms of performance.  For the split-merge algorithm
\cite{jain04splitmerge}, we only use the third variant that performs
intermediate Gibbs sampling.  The other split-merge algorithms fared
surprisingly poorly---often losing significantly even to standard
Gibbs in terms of log likelihood.

The results of these experiments are shown in
Figure~\ref{fig:artificial}.  The upper-left graph shows a plot of the
data negative log likelihood as a function of data set size (lower is
better).  The negative log likelihoods are presented as a ratio to the
log likelihood of the true MAP assignment.  Interestingly, \emph{both}
inadmissible search algorithms achieve optimal performance.  The
beam-based admissible heuristics do worse, and the Gibbs sampler never
reaches optimal performance (except on the smallest data set).  The
split-merge performance is quite variable: sometimes better than
Gibbs, sometimes worse.

In the upper-right graph, we plot computation time as a function of
data set size.  Split-merge is almost always slightly slower than
Gibbs.  As can be easily seen, the time taken for both admissible
heuristics grows quite fast and, at least for the trivial heuristic,
becomes intractable after only 10-15 data points.  The Trivial-Beam
remains reasonably efficient (on par with Gibbs), but also had the
worst negative log likelihood performance.  The Admissible-Beam is
actually reasonably slow because the computation of the heuristic is
time-consuming.  As expected, the inadmissible heuristic (with or
without beam) is always the fastest.

The bottom-left graph shows the f-score (geometric mean of precision
and recall) of pair-wise ``same cluster or not'' decisions for the
clustering found by the different algorithms against the ground truth
(higher is better).  As we can see, the inadmissible heuristic always
performs best, while the others vary significantly in performance
(Gibbs, for some reason, does quite poorly initially but then improves
as the size of the data set increases).

Finally, the bottom-right graph shows (for only the search-based
algorithms) the number of states enqueued during search.  This
corresponds roughly with the upper-right graph (time) but excludes
considerations such as the time to compute the heuristic.  As
expected, the inadmissible heuristic enqueues by far the fewest
entries (in fact, for most of these algorithms, the number of elements
\emph{dequeued} by the inadmissible heuristic was between $N$ and
$N+5$ for all algorithms, meaning that pure greedy search may be
reasonable).

\subsection{HANDWRITTEN DATA}

For this experiment, we use the handwritten data set from MNIST
(specifically, the version assembled by Sam Roweis) consisting of
images of numbers (in $28 \times 28 = 784$ dimensions).  Following
\namecite{kurihara06accelerated}, we preprocess the data by centering
and spherizing it, then running PCA to obtain a $50$-dimensional
representation.  This data set consists of $60,000$ images, and thus
only the inadmissible heuristic is sufficiently efficient to run (we
use a beam of 100, though this turns out to be unnecessarily large).
We run on three versions of the data: a $5\%$ subset, a $20\%$ subset
and the full set.  In all cases, we use $\al = 1$ and a prior variance
of $0.1$.

At the $5\%$ level, search completes in just over $11$ seconds,
roughly $270$ data points per second.  The algorithm finds a solution
using $11$ clusters and achieves a negative log likelihood of
$2.04e5$.  Running Gibbs on this data set takes roughly $40$ seconds
\emph{per iteration}, and after $100$ iterations achieves a best
performance of $2.09e5$.  Split-merge obtains a log-likelihood of
$2.05e5$, but takes almost two hours to do so.  At the $20\%$ level,
search completes in just over $105$ seconds ($115$ elements per
second) with a final negative log likelihood of $8.02e5$ (and 21
clusters).  Gibbs sampling takes roughly $18$ \emph{minutes} per
iteration and find a best solution of $8.34e5$ (after $100$
iterations).  Split-merge obtains a solution with negative log
likelihood of $8.15e5$ after eight hours.  Finally, for the full data
set, search completes in just under $15$ minutes (roughly $66$ data
points per second) with a score of $3.96e6$ (and 27 clusters).  Gibbs
takes an egregious $7$ hours per iteration on this data, so we were
only able to run $15$ iterations, leading to a best score of $4.2e6$.
Split-merge was also to slow to run for more than $15$ iterations,
achieving a score of $4.1e6$ after about $7$ days.

\begin{figure}[t]
\center
\psfig{figure=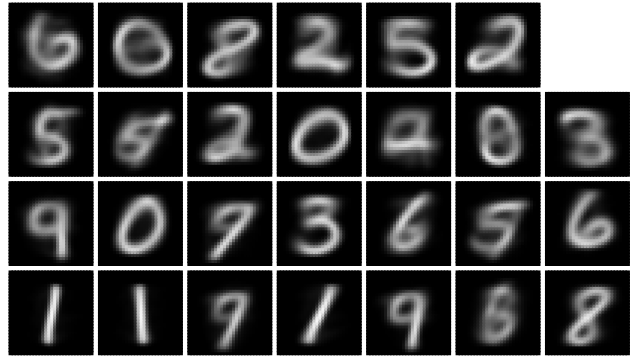,width=8.3cm}
\caption{Mean images for clusters found on the full ($60k$ examples)
MNIST data set.}
\label{fig:images}
\end{figure}

In Figure~\ref{fig:images}, we show the mean image from each cluster
found by the search algorithm on the full data set (sorted by cluster
size).  Qualitatively, these clusters appear quite reasonable.
(\namecite{kurihara06accelerated} present a similar figure for
variational techniques; however, their model uses a full
Gaussian/inverse-Wishart prior while ours only uses a Gaussian and we
use a fixed prior variance.  One could easily adapt our algorithms to
use the full prior, but we do not do so in the experiments reported
here.)

\subsection{NIPS DOCUMENTS} \label{sec:nips}

Finally, to demonstrate the applicability of our algorithm to discrete
data, we apply the same model to the set of papers from NIPS 1--12
(assembled by Sam Roweis).  This data set consists of $1740$ documents
over a vocabulary of roughly $13k$ words.  In our model, we drop the
top ten words from the vocabulary and retain only the top 1000 from
the remainder.  For this data, we use the Dirichlet/Multinomial DP
mixture model with $\al = 1$ and a symmetric Dirichlet prior with
parameter equal to $10$.

\begin{table}[t]
\begin{center}
\begin{small}
\begin{tabular}{|@{~}p{3.2in}@{ }|}
\hline
 \textsf{training hidden units error learning weight generalization network weights regression layer algorithm recurrent gradient nodes prediction theorem convergence node student} \\
\hline
 \textsf{spike neuron neurons cells synaptic firing cell cortical cortex activity synapses stimulus excitatory orientation inhibitory membrane spikes ocular dominance fig} \\
\hline
 \textsf{mixture speech em likelihood image tangent hmm clustering word images pca recognition posterior bayesian speaker gaussian experts kernel cluster classification} \\
\hline
 \textsf{chip circuit analog motion image voltage vlsi visual auditory images velocity sound retina intensity optical pulse disparity template pixel silicon} \\
\hline
 \textsf{policy reinforcement state controller action control actions robot reward agent mdp learning states sutton policies trajectory planning singh barto rl} \\
\hline
 \textsf{object objects units views face image hidden unit visual images recognition attractor activation faces features layer attention feature representations module} \\
\hline
 \textsf{bounds threshold polynomial bound theorem depth boolean proof gates lemma dimension tree node concept clause class functions winnow boosting maass} \\
\hline
 \textsf{motion head motor visual eeg eye direction subjects stimulus ica movements cells movement cortex velocity cue field parietal vor spatial} \\
\hline
 \textsf{word classifiers classifier hmm rbf character recognition training speech mlp characters hybrid user context net hidden layer trained words error} \\
\hline
 \textsf{belief evidence similarity posterior bayesian retrieval user propagation hypotheses query concept approximation examples exact images iii image mackay database nodes} \\
\hline
\end{tabular}
\end{small}
\end{center}
\caption{Most indicative words for each of the ten largest clusters
  (out of fourteen) of documents from NIPS papers.  Sorted by cluster
  size.}
\label{tab:words}
\end{table}

Running DP search (with the inadmissible heuristic) on this data set
takes roughly $21$ seconds ($83$ documents per second) and results in
fourteen clusters.  For each cluster, we extract the top twenty words
from the documents assigned to that cluster (``top'' as determined by
tf-idf score).  These are depicted in Table~\ref{tab:words} (sorted by
cluster size).

We also experiment with alternative orderings of the data set for this
problem.  When the data is presented in \emph{ascending} order of
marginal likelihood, the resulting log likelihood is $2.4407e6$.  When
this order is reversed, the resulting log likelihood is $2.4740e6$.
Finally, we consider presentations in random order.  Over ten such
orders, the mean log likelihood is $2.4489e6$ and the variance is
$0.0067e6$.  (Of all the random passes, only one achieves a higher log
likelihood than the default ascending order, and does so with a small
gain: $2.4403e6$.)  This suggests that the ascending order is a
reasonable heuristic.

In comparison the vanilla Gibbs sampling and the split-merge
proposals, the search algorithm again performs significantly better.
The log likelihood for the best Gibbs clustering on this data was
$3.2e6$ and for the split-merge proposals it as $3.0e6$, both taking a
little over an hour.

\section{PRIOR WORK} \label{sec:priorwork}

Although in this paper we have only compared our search-based
technique to straightforward Gibbs sampling, there are other inference
techniques for DP mixture models.  Still within the context of MCMC,
\namecite{xing04haplotype} propose a Metropolis-Hastings sampler that
is shown to mix faster than the Gibbs sampler and is only moderately
more challenging to implement.  An alternative, recent proposal for
inference in DP mixture models is to make use of particle filters
(sequential MCMC) \cite{fearnhead04particledp}.  Particle filters look
somewhat like a stochastic beam search algorithm and, as such, are
similar in spirit to the approach proposed here.

We are additionally aware of two deterministic approaches to inference
in DP mixture models based on variational techniques.  The first, due
to \namecite{blei05variational}, employs the stick-breaking
construction for the Dirichlet process \cite{sethuraman-SS1994} to
construct a finite variational distribution for the infinite mixture
model.  On artificial data, they report on the order of a $100-300$
decrease in time (over Gibbs sampling) with essentially identical
\emph{held-out} data log likelihoods.  Very recently,
\namecite{kurihara06accelerated} present even more efficient
variational algorithm for DP mixture models.  In contrast to the
method of \namecite{blei05variational}, the new algorithm employs an
infinite variational distribution and a collapsed distribution
\cite{kurihara07collapsed}.  It can be further accelerated (at least
in the Gaussian case) by using kd-trees for caching sufficient
statistics of the data set.

\section{CONCLUSIONS}

We have presented an algorithm for finding the MAP clustering for data
under a Dirichlet Process mixture model, a task that appears regularly
in many practical applications.  It has been shown to be extremely
efficient (clustering a data set of $60k$ elements in under $15$
minutes in Matlab) and general (we have shown applications both to
continuous and discrete data sets).  Moreover, for small data sets, we
have shown relatively efficient schemes for finding provably optimal
solutions to the MAP problem.  Our results, especially with the
inadmissible scoring function, show that one can obtain a very good
approximate MAP solution incredibly quickly.  Profiling shows that our
main bottleneck is actually optimizing $p(\vec c)$ from
Section~\ref{sec:c}, not $p(\vec x \| \vec c)$.  In the worst case,
this optimization is \emph{quadratic} in the size of the data set, not
linear.  We are currently investigating ways of making this more
efficient.

In comparison to variational approaches to DP mixture models
\cite{blei05variational,kurihara06accelerated}, our algorithm is
applicable to exactly the same data (exponential family with conjugate
priors) and suffers from the same drawback: one cannot easily
re-estimate the concentration parameter.  However, given the speed of
our algorithm, one could easily use multiple runs with Bayesian model
selection to find a suitable value.  It should be noted that the
results presented in papers discussing variational approaches compare
to Gibbs sampling in terms of log likelihood and speed.  The general
result is that the variational approaches obtain similar log
likelihoods about $100-300$ times faster.  Our results with the
inadmissible function show that we can often achieve \emph{better}
results to Gibbs.  It is difficult to imagine an algorithm that is
more computationally efficient than search with our inadmissible
function.

The primary advantage of MCMC techniques over our method is that they
produce a true representation of the posterior, provided that they are
run for long enough\footnote{As we observed in
Section~\ref{sec:artificial}, the Gibbs sampler often fails to ever
reach the MAP solution.}  However, if a true sample of the posterior
is desired, it would be natural to run our algorithm as an initializer
for any MCMC algorithm.  This would yield the benefits of a sample
from the posterior without requiring that the sampler first find a
region of high posterior probability.  Additionally, MCMC techniques
\emph{can} be applied to non-conjugate distributions (at least in
theory) by using an embedded sampling procedure to estimate the
intractable integrals.  One could, in principle, use the same
methodology within our search framework, though this would obviate
many of the speed benefits.  An alternative would be to use an
efficient deterministic approximation.

\paragraph{Acknowledgments.}  Thanks to Andy Carlson and the
three anonymous reviewers whose constructive criticism and pointers
helped improve this paper.

\bibliographystyle{fullname}
\bibliography{bibfile}

\end{document}